\documentclass[journal]{IEEEtran}
\usepackage{graphicx}
\ifCLASSINFOpdf
\else
\fi
\usepackage{booktabs}
\usepackage{amsmath}
\usepackage{amssymb}
\usepackage{multirow}
\usepackage{array}
\usepackage{algorithm}
\usepackage{algorithmic}
\ifCLASSOPTIONcompsoc
 \usepackage[caption=false,font=normalsize,labelfont=sf,textfont=sf]{subfig}
\else
 \usepackage[caption=false,font=footnotesize]{subfig}
\fi
\begin{document}
%
% paper title
% Titles are generally capitalized except for words such as a, an, and, as,
% at, but, by, for, in, nor, of, on, or, the, to and up, which are usually
% not capitalized unless they are the first or last word of the title.
% Linebreaks \\ can be used within to get better formatting as desired.
% Do not put math or special symbols in the title.
\title{FPB: Feature Pyramid Branch for Person Re-Identification}
%
%
% author names and IEEE memberships
% note positions of commas and nonbreaking spaces ( ~ ) LaTeX will not break
% a structure at a ~ so this keeps an author's name from being broken across
% two lines.
% use \thanks{} to gain access to the first footnote area
% a separate \thanks must be used for each paragraph as LaTeX2e's \thanks
% was not built to handle multiple paragraphs
%

\author{Suofei Zhang,
       Zirui Yin,
       Xiofu Wu,
       Kun Wang,
       Quan Zhou 
       and~Bin~Kang% <-this % stops a space
% \thanks{This work was jointly supported in part by the National Natural Science Foundation of China under Grants 61876093 and 61801242, the National Natural Science Foundation of Jiangsu Province under Grant BK20181393.}
% \thanks{Suofei Zhang, and Bin Kang are with the School of Internet of Things, Nanjing University of Posts and Telecommunications, Nanjing 210003, China (e-mails: zhangsuofei@njupt.edu.cn; kangb@njupt.edu.cn).}% <-this % stops a space
% \thanks{Kun Wang is with the Department of Electrical and Computer Engineering, University of California, Los Angeles, CA, USA (e-mail:wangk@ucla.edu).}
% \thanks{Zirui Yin, Xiofu Wu and Quan Zhou are with the National Engineering Research Center of Communications and Networking, Nanjing University of Posts and Telecommunications, Nanjing 210003, China (e-mails: 1019010621@njupt.edu.cn; xfuwu@ieee.org; quan.zhou@njupt.edu.cn).}% <-this % stops a space
}%\thanks{Manuscript received April 19, 2005; revised August 26, 2015.}

% note the % following the last \IEEEmembership and also \thanks - 
% these prevent an unwanted space from occurring between the last author name
% and the end of the author line. i.e., if you had this:
% 
% \author{....lastname \thanks{...} \thanks{...} }
%                     ^------------^------------^----Do not want these spaces!
%
% a space would be appended to the last name and could cause every name on that
% line to be shifted left slightly. This is one of those "LaTeX things". For
% instance, "\textbf{A} \textbf{B}" will typeset as "A B" not "AB". To get
% "AB" then you have to do: "\textbf{A}\textbf{B}"
% \thanks is no different in this regard, so shield the last } of each \thanks
% that ends a line with a % and do not let a space in before the next \thanks.
% Spaces after \IEEEmembership other than the last one are OK (and needed) as
% you are supposed to have spaces between the names. For what it is worth,
% this is a minor point as most people would not even notice if the said evil
% space somehow managed to creep in.

% The paper headers
\markboth{Journal of \LaTeX\ Class Files,~Vol.~14, No.~8, August~2015}%
{Shell \MakeLowercase{\textit{et al.}}: Bare Demo of IEEEtran.cls for IEEE Journals}
% The only time the second header will appear is for the odd numbered pages
% after the title page when using the twoside option.
% 
% *** Note that you probably will NOT want to include the author's ***
% *** name in the headers of peer review papers.                   ***
% You can use \ifCLASSOPTIONpeerreview for conditional compilation here if
% you desire.

% If you want to put a publisher's ID mark on the page you can do it like
% this:
%\IEEEpubid{0000--0000/00\$00.00~\copyright~2015 IEEE}
% Remember, if you use this you must call \IEEEpubidadjcol in the second
% column for its text to clear the IEEEpubid mark.

% use for special paper notices
%\IEEEspecialpapernotice{(Invited Paper)}

% make the title area
\maketitle

% As a general rule, do not put math, special symbols or citations
% in the abstract or keywords.
\begin{abstract}  
  High performance person Re-Identification (Re-ID) requires the model to focus on both global silhouette and local details of pedestrian.
  To extract such more representative features, an effective way is to exploit deep models with multiple branches.
  However, most multi-branch based methods implemented by duplication of part backbone structure normally lead to severe increase of computational cost.
  In this paper, we propose a lightweight Feature Pyramid Branch (FPB) to extract features from different layers of networks and aggregate them in a bidirectional pyramid structure.
  Cooperated by attention modules and our proposed cross orthogonality regularization, FPB significantly prompts the performance of backbone network by only introducing less than 1.5M extra parameters.
  Extensive experimental results on standard benchmark datasets demonstrate that our proposed FPB based model outperforms state-of-the-art methods with obvious margin as well as much less model complexity.
  FPB borrows the idea of the Feature Pyramid Network (FPN) from prevailing object detection methods.
  To our best knowledge, it is the first successful application of similar structure in person Re-ID tasks, which empirically proves that pyramid network as affiliated branch could be a potential structure in related feature embedding models.
  The source code is publicly available at https://github.com/anocodetest1/FPB.git.
\end{abstract}

% Note that keywords are not normally used for peerreview papers.

% For peer review papers, you can put extra information on the cover
% page as needed:
% \ifCLASSOPTIONpeerreview
% \begin{center} \bfseries EDICS Category: 3-BBND \end{center}
% \fi
%
% For peerreview papers, this IEEEtran command inserts a page break and
% creates the second title. It will be ignored for other modes.
\IEEEpeerreviewmaketitle

\section{Introduction}\label{sec:introduction}
\IEEEPARstart{P}{erson} Re-Identification (Re-ID) has been extensively applied as a retrieval technique in large scale person tracking and related scenarios of intelligent video surveillance.
Given a query sample of specific person, it aims to match the same person in the gallery set of samples which may be captured by cameras from different viewpoints in different backgrounds~\cite{zheng2016person,9336268}.
Due to the increasing demand of public safety, the practical importance of person Re-ID leads to more and more attention from community.
Although significant progress has been witnessed during the last decade, there still exist some challenging problems in the study of person Re-ID, e.g., partial occlusions ~\cite{8578633,8954276}, drastic deformations of human pose~\cite{8099586,8578149}, complex environment and background clutter~\cite{8578227}, etc.
\begin{figure}
  % \vskip 0.2in
  \centering
  \includegraphics[width=0.5\textwidth]{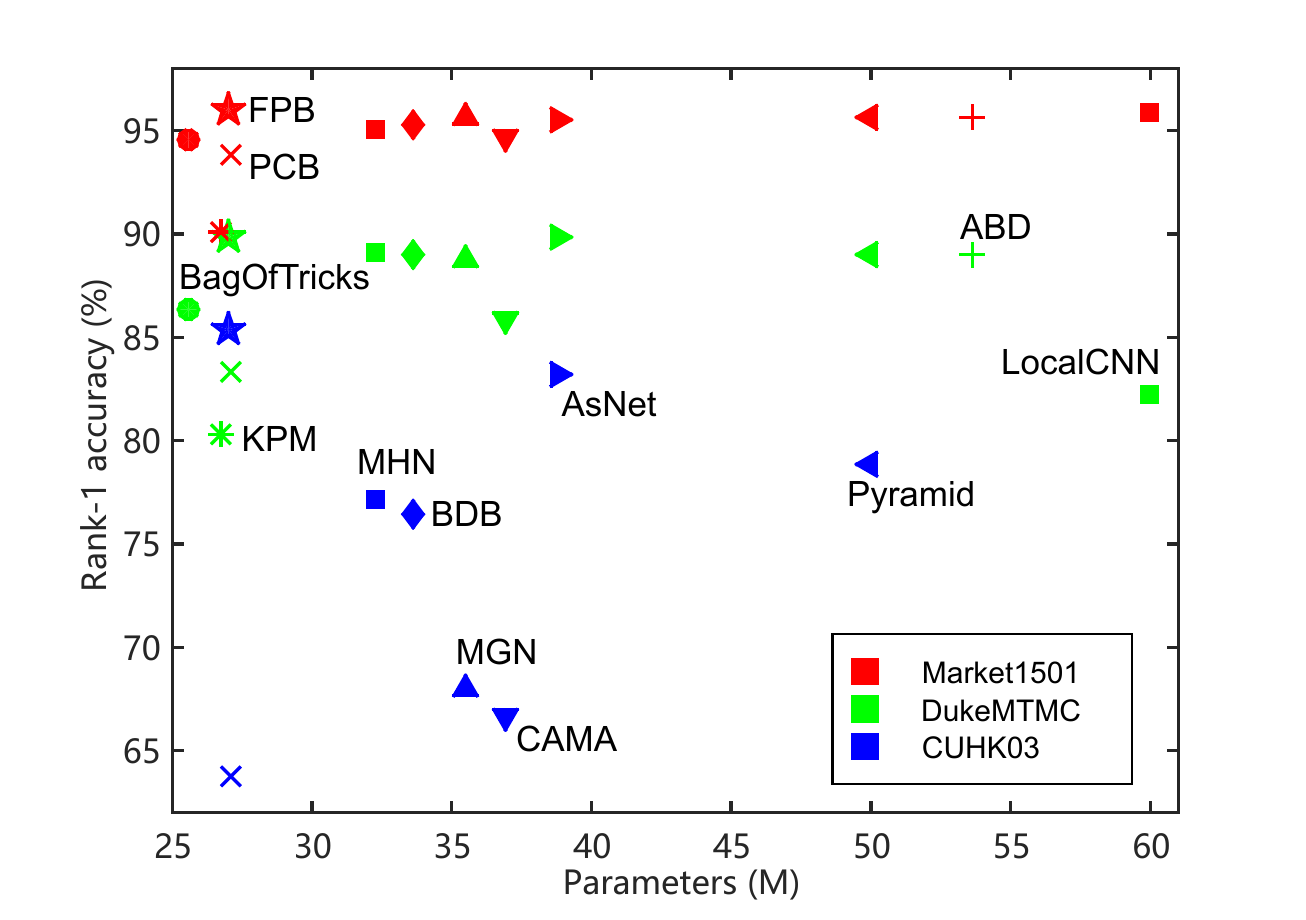} \\
  \includegraphics[width=0.5\textwidth]{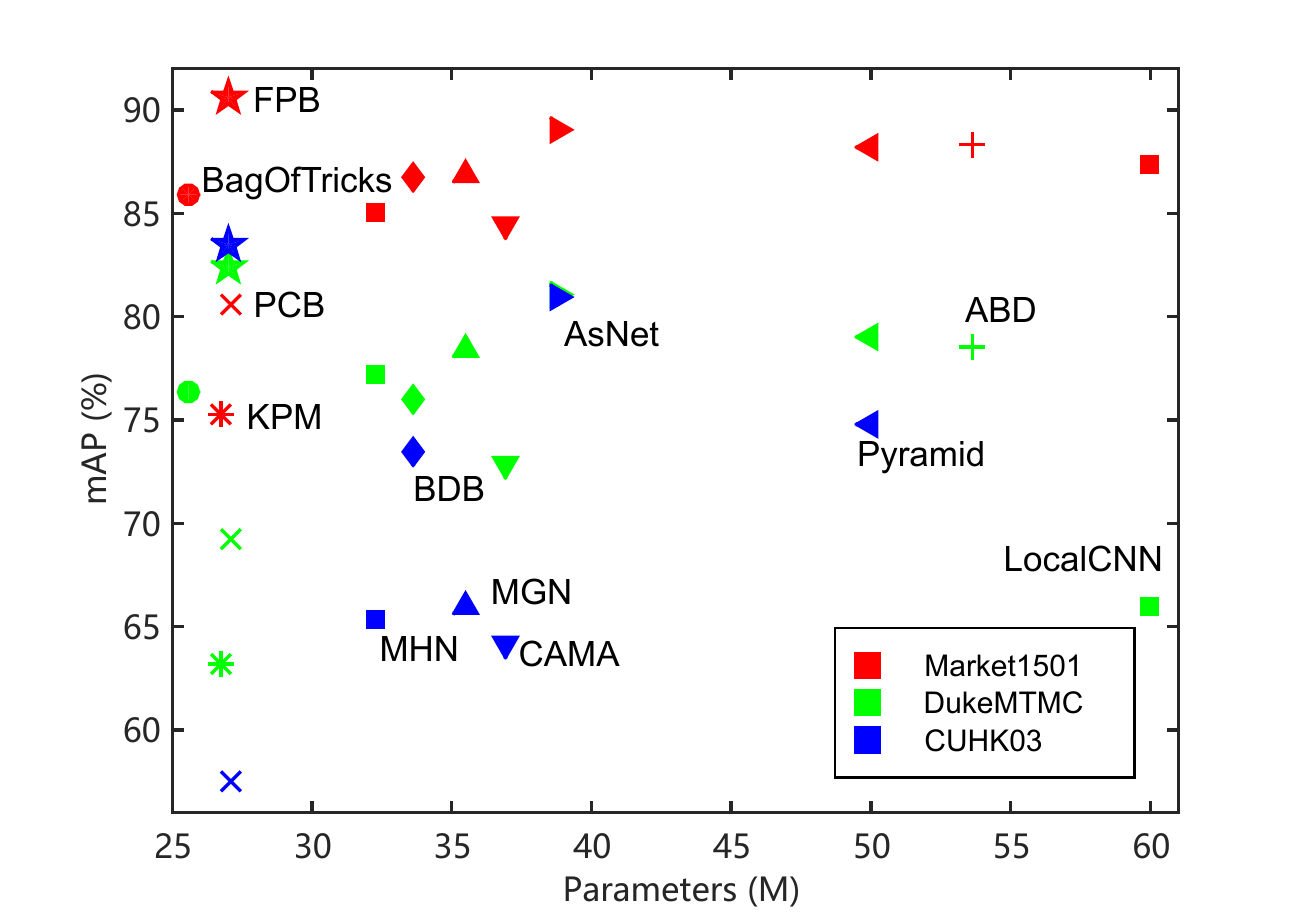}
  % \subfloat[]{\includegraphics[width=0.5\textwidth]{maps.pdf}}\label{fig:maps_comp} \\
  % \subfloat[]{\includegraphics[width=0.5\textwidth]{rank1.pdf}}\label{fig:r1_comp}
  \caption{Performance comparison of our proposed FPB and other state-of-the-art Re-ID methods. 
  Results on the same dataset are presented by the same color.
  Results of the same method are presented by the same shape.
  Top: mAP vs. number of model parameters.
  Bottom: Rank-1 accuracy vs. number of model parameters.
  }
  \label{fig:maps}
  % \vskip -0.2in
  \end{figure}

Recently, deep learning based models have been proven to be quite effective to tackle aforementioned problems~\cite{zheng2016person,sun2018beyond,luo2019bag}.
Pre-trained Convolutional Neural Network (CNN) models such as ResNet~\cite{he2016deep} and InceptionNet~\cite{szegedy2017inception} serve as a strong backbone for extracting representative visual features from images.
The generic framework of such approaches mainly follows a fine-tuning stage on labeled samples to deliver a model which can discriminate person from each other within training set.
Then intermediate features before final classification layer are retained to accomplish the deep learning based \textit{feature embedding}.
These low-dimensional but representative features instead of original images of person can be exploited for high efficient matching as well as retrieval of unknown person in new set of instances.

In contrast to conventional vision tasks with large scale datasets, existing datasets for person Re-ID normally require algorithm to learn a classification model over a relatively large number of classes ($>700$ persons) with limited samples ($<20,000$ images).
Also, due to the slight difference between the final goal of feature embedding and learning of classification, simply using existing backbone models in Re-ID tends to deliver intermediate features without sufficient generalization.
Many efforts have been devoted to alleviate this problem, including 1) particularly designed loss function to discover discriminative features, e.g., sphere loss~\cite{fan2019spherereid}, triplet loss~\cite{hermans2017defense} and center loss~\cite{wen2016discriminative}; 2) adopting affiliated structure such as local part~\cite{sun2018beyond,9124699} or multiple branches~\cite{yang2019CAMA,zheng2019pyramidal,9351775} to learn fine-grained features for higher diversity; 3) introducing attention mechanisms to emphasize feature correlations and prompt the efficiency of model~\cite{chen2019ABD}.
However, recent works focus on tediously adding extra structures to backbone networks for the gain of performance, which raises an interesting question.
Are these structures really effective or performance can also be prompted by simply increasing the number of parameters, e.g., model with larger backbones or ensemble techniques~\cite{paisitkriangkrai2015learning}.

In this paper, we comprehensively consider the performance as well as efficiency of person Re-ID models, proposing a compact model consisting of a backbone and a Feature Pyramid Branch (FPB).
FPB is mainly inspired by the Feature Pyramid Network (FPN) structure in the field of object detection.
As a common structure in prevailing object detectors~\cite{8099589,9156454,bochkovskiy2020yolov4}, FPN has been proven to be quite effective in aggregating features at different scales.
Although many works in the literature have proven that person Re-ID also requires feature extraction at different granularities~\cite{sun2018beyond,wang2018parameter}, due to the difference between tasks of Re-ID and object detection, it is still challenging to exploit FPN into person Re-ID architecture.

Here we address this problem, proposing a bidirectional pyramid structure cooperated by attentive auxiliary modules as a lightweight branch solution.
In Fig.~\ref{fig:maps}, we compare our proposed model with other ResNet based state-of-the-art methods.
We choose ResNet50 as a typical backbone instance since it is extensively adopted by existing Re-ID methods and shows high compatibility with various structures.
Based on ResNet50, a backbone with 25.56M parameters, our proposed FPB based method achieves the best performance on all benchmarks by only introducing less than 1.5M extra parameters.

From Fig.~\ref{fig:maps} one can see that, 1) for methods with parallel amounts of parameters as FPB ($<$28M), e.g., Bag-Of-Tricks, PCB and KPM, the mean Average Precision (mAP) is significantly prompted over 5\% on all of three datasets.
2) Other leading methods with approaching performance normally require models with more than 32M parameters, which implies at least 4$\times$ times extra parameters are added to the backbone.
3) Furthermore, FPB also outperforms methods with much higher complexity such as Pyramid, ABD and LocalCNN, which retain models with more than 50M learnable parameters.
Our main contributions in this paper can be summarized as follows:
\begin{enumerate}
  \item We propose a lightweight FPB which can be plugged into backbone network to form an asymmetrical multi-branch architecture.
  Diverse features from different scales are extracted and integrated for final matching.
  As far as we know, we are the first to successfully exploit feature pyramid network into the model of person Re-ID.
  % By a careful trade-off between model capacity and efficiency, FPB significantly prompt the final performance by only introducing less than 1.5M extra parameters.
  \item To further prompt the performance of FPB, self-attention modules as auxiliary modules are carefully evaluated and inserted into different positions of network.
  Also, we propose an extra cross orthogonality regularization over features from two layers of FPB.
  This penalty effectively reduces the correlation of feature maps especially after attention modules, and thus improves the diversity of resulting features.
  \item Extensive validations on different person Re-ID datasets demonstrate that FPB structure can deliver significant improvement with trivial increase of computation cost ($<$1.5M extra parameters).
  It outperforms other leading methods and achieves new state-of-the-art on all prevailing benchmarks with an efficient implementation.
  Our results also empirically prove that FPN could be a potential structure in related feature embedding tasks.
\end{enumerate}

\section{Related works}\label{sec:related-works}
\subsection{Person Re-Identification}
Recently, deep learning based person Re-ID methods show clear advantages to conventional methods which still rely on handcrafted features~\cite{khamis2014joint,6247939}.
There also exist two paradigms to learn deep models with the feature embedding capability for final matching.
The first one is to adopt pre-trained backbone, e.g., ResNet, and fine-tune parameters of specific architecture on person Re-ID datasets~\cite{sun2018beyond,luo2019bag,yang2019CAMA,zheng2019pyramidal,chen2019ABD,9094042}.
The other one is to design novel architecture specifically for the task of person Re-ID and train the model from scratch~\cite{8578341,9011001,lawen2019attention,10.1007/978-3-030-60636-7_2}.
Although specific models such as Omni-Scale Network (OSNet)~\cite{9011001} and Harmonious Attention CNN (HA-CNN)~\cite{8578341} normally imply higher efficient models with much less parameters, we still focus on the former paradigm based on generic backbones in this paper.
This is because we observe that these models deliver higher performance especially on large scale benchmarks, e.g., MSMT17~\cite{8578114}.
Also, they demonstrate higher feasibility for further modification as well as distribution in realistic scenarios.

On the other hand, different mechanisms are also introduced into the stage of model training to prompt the performance.
Data augmentation methods, such as random erasing~\cite{zhong2020random}, random patch~\cite{9011001}, are commonly adopted by various methods.
Based on essential random erasing at every single sample, randomly dropping block~\cite{dai2019BDB, wu2020diversityachieving} further proposed to randomly drop the same part at all samples within a batch for learning more attentive local features.
Besides augmentation, other strategies such as stochastic weight averaging~\cite{izmailov2018averaging}, extra regularization~\cite{ni2020adaptive} have also been proven to be effective to prompt the capability of learned models.
In this paper, since we focus on the structure of FPB, only essential strategies during training are adopted for a fair comparison.

\subsection{Diversity of Features}
Due to aforementioned reasons, to prompt the generalization capacity of feature embedding models is a critical issue in person Re-ID.
An effective way is to increase the diversity of extracted features.
Conventional output feature of deep models is normally resulted by direct averaging operation over the whole feature map.
Conversely, \cite{sun2018beyond} proposed to extract local features from different parts for higher diversity in final matching.
Besides part based models, to extract features from multiple levels~\cite{8578346,8237839} is another potential way to deliver more representative features.
Features from shallow layers of networks can naturally reflect detailed local information at images.
On the other side, recent research demonstrates that a complementary feature consisting of both global and local features could be more diverse for feature matching.
Thus multi-branch structure becomes a prevailing architecture in the literature of person Re-ID~\cite{chen2019ABD,yang2019CAMA,10.1007/978-3-030-60636-7_2,zheng2019pyramidal}.

An inevitable problem brought by multi-branch structure is the increase of complexity.
As an essential way, \cite{chen2019ABD,9094042} construct the dual-branch structure by duplicating part of backbone with separated learnable parameters.
For more branches, since most of existing works tend to design symmetrical multi-branch structure, multiple duplications of the last part of backbone apparently lead to significant increase of parameters. 
For some lightweight backbones, multiple branches could even introduce extra parameters more than original backbones~\cite{lawen2019attention,10.1007/978-3-030-60636-7_2}.
Our proposed approach follows the paradigm of multi-branch networks.
However, we carefully restrain the complexity of local part branch by a lightweight pyramid network structure rather than direct duplication of backbone.
The pyramid network can also extract features from multiple levels of backbone, thus detailed information from shallow layers are naturally retained.
Finally, complementary feature embedding at different granularities are learned jointly in an end-to-end framework.

\subsection{Feature Pyramid Network}
The FPN structure is extensively exploited in the field of object detection.
The first successful application of FPN is~\cite{8099589}.
It proposed a top-down pathway to output detection results at multiple scales simultaneously.
In this way, conventional pyramid of original input image is replaced by a CNN structure to achieve multi-scale object detection more efficiently.
Following the idea, PANet~\cite{8579011} further proposed a bidirectional FPN consisting of a top-down as well as a bottom-up path to aggregate features at each layer of FPN.
M2det~\cite{zhao2019m2det} adopted a block of alternating joint U-shape module to fuse multi level features.
EfficientDet~\cite{9156454} introduced the down-sampling structure from ResNet and proposed a Bidirectional FPN BiFPN.
State-of-the-art object detectors including both one-stage approaches (SSD~\cite{978-3-319-46448-02}, YOLO~\cite{bochkovskiy2020yolov4}, EfficientDet~\cite{9156454}) and two-stage approaches (Mask RCNN~\cite{he2017mask}, DetNet~\cite{Li_2018_ECCV}) all exploit FPN structure to tackle the scale variation problem. 
In this paper, we introduce the FPN into the task of person Re-ID since it also requires feature matching at both global and local scales.
Despite difference exists between person Re-ID and object detection, our results empirically prove that with delicate design, FPN can bring significant benefit with trivial increase of model complexity.

\subsection{Attention Mechanisms}
There are various implementations of attention modules in the literature.
The original attention module is proposed for the Natural Language Processing (NLP) tasks~\cite{bahdanau2014neural,vaswani2017attention}, which is normally referred as the multi-head attention.
It focuses on reducing the ambiguity of input features by their context information.
Recently, multi-head attention as well as the transformer architecture have also been proven to be effective in various vision tasks~\cite{dosovitskiy2020image}.
Differing from original multi-head attention module in NLP, the most extensively applied attention module in vision tasks is the so-called Non-Local Neural Networks structure~\cite{8578911}.
It aims to encode the correlation between features at different positions to output more attentive features as well.
Differing from Non-Local Neural Networks, another design of attention, namely Position Attention Module (PAM) is also extensively adopted in person Re-ID tasks~\cite{chen2019ABD}.
PAM can be treated as a simplified version of multi-head attention without dimension reduction or multi-head structure on the value branch.
It also focuses on attending features at different positions, extracting correlation information as position affinity matrix to reweigh features.

Besides PAM, there exists another series of attention modules, namely Channel Attention Module (CAM), aim to extract correlation information over different channels of feature maps.
Some typical implementations of CAM can be referred to Squeeze-and-Excitation block~\cite{8701503} and Efficient Channel Attention~\cite{wang2019ecanet}.
Since no extra parameter is required in the implementation of CAM, it can be deployed as an efficient mechanism to extract channel-wise response over features with trivial cost. 
Many variants of CAM attend as building block in different architectures of CNNs, specifically in lightweight designs such as OSNet~\cite{9011001} and EfficientNet~\cite{tan2019efficientnet}.
Also, CAM is extensively deployed in existing person Re-ID frameworks such as HA-CNN~\cite{8578341} and Attentive but Diverse Network (ABD-Net)~\cite{chen2019ABD}.

In this paper, our implementation of attention module consists of the concatenation of a PAM and a CAM.
This structure is analogous to the design of attention module in \cite{chen2019ABD}.
We observe that the PAM can deliver slightly better performance than prevailing Non-Local Neural Networks in our architecture.
Meanwhile, the CAM can also bring explicit gain via negligible increase of parameters. 

\section{Proposed Method}
\begin{figure*}
  % \vskip 0.2in
  \begin{center}
  \centerline{\includegraphics[width=0.9\textwidth]{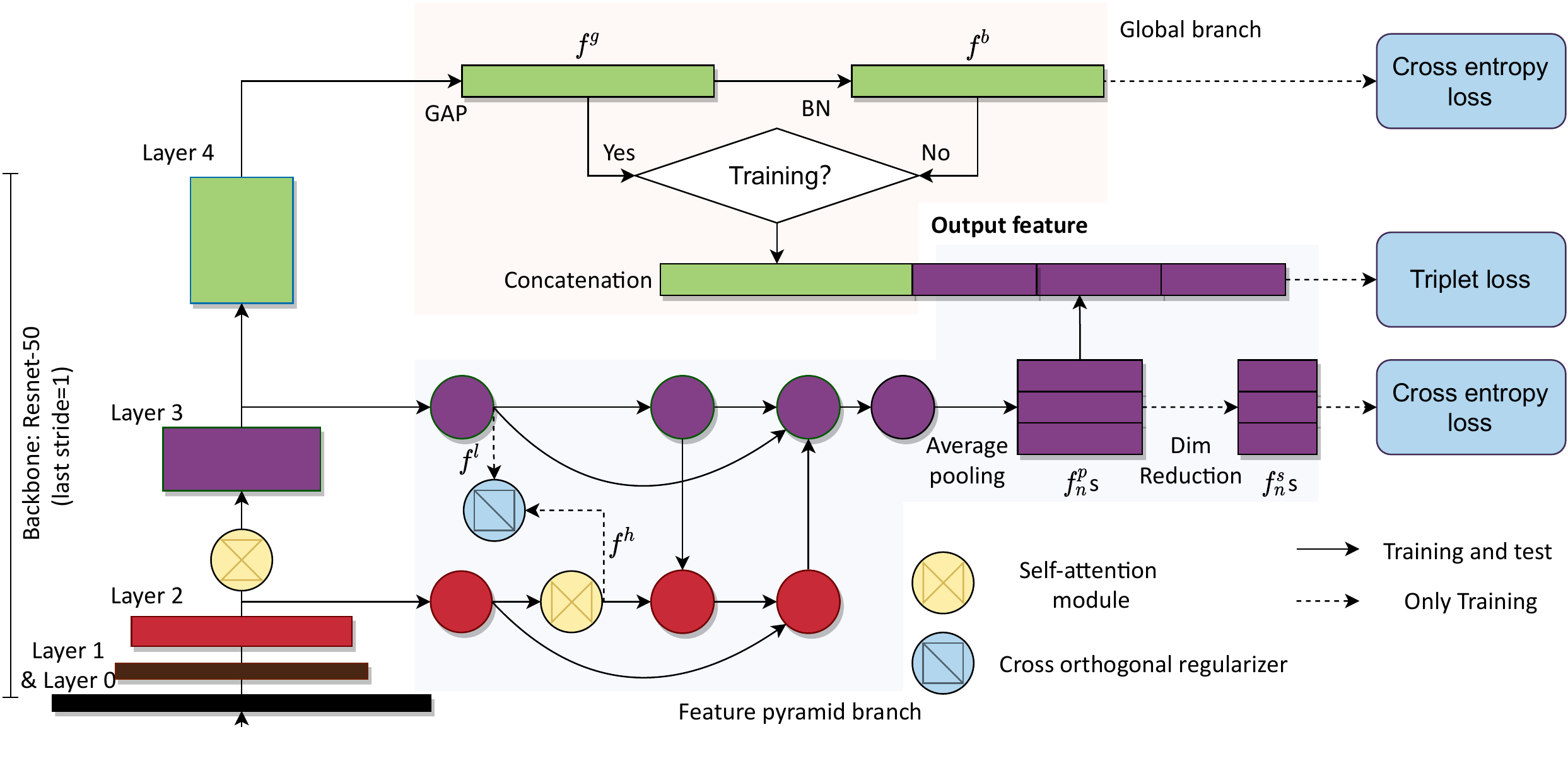}}
  \caption{The overall architecture of our proposed method.
  Here the stride of the last down-sampling layer is changed from 2 to 1 and denoted as last stride=1.
  The residual structure in FPB is illustrated as curved arrows here.}
  \label{fig:architecture}
  \end{center}
  % \vskip -0.2in
  \end{figure*}

As illustrated in Fig.~\ref{fig:architecture}, our proposed method is a dual-branch framework, consisting of a \textit{global branch} and a \textit{feature pyramid branch}.
The \textit{global branch} is mainly based on the Bag-Of-Tricks method~\cite{luo2019bag}, which contains a modified version of ResNet50~\cite{he2016deep} as backbone network.
Differing from standard ResNet50 based ID-Discriminative Embedding (IDE)~\cite{zheng2016person}, here the last down-sampling operation in layer 4 of ResNet50 is removed to increase the size of the output.
After the Global Average Pooling (GAP) operation over the output, a 2048-dimension vector is delivered as the global feature $\boldsymbol{f}^g$.
A BNNeck~\cite{luo2019bag} is also adopted to diliver the normalized version of global feature $\boldsymbol{f}^b$ here.
More details can be referred to the description of loss fuctions in Section~\ref{sec:loss}.

Based on the backbone network, we propose a lightweight \textit{Feature Pyramid Branch} (FPB) to enrich the diversity of features for person Re-ID.
The outter structure of this branch is inspired by the Part-based Convolutional Baseline (PCB)~\cite{sun2018beyond}.
Output feature map is devided into parts to emphasize local information.
The difference is we take shallower features from the layer 2 and layer 3 of backbone as input here.
These shallow features can retain more local details from image.
Simultaneously, features with lower dimensions reduce the number of learnable parameters within the branch.

Specificially, the output feature map of FPB is processed by different strategies during training and inference respectively.
Average pooling operation is adopted here to archive $N$ 1024-dimension vectors as $\boldsymbol{f}_n^p$s ($n\in\{1,\ldots,N\}$).
A Dimension Reduction (DM) layer consisting of convolutional filter followed by Batch Normalization (BN) and Rectified Linear Unit (ReLU) activation compresses $\boldsymbol{f}_n^p$s to $N$ 256-dimension vectors as $\boldsymbol{f}_n^s$s to optimize the classification loss on the branch during training.
The concatenation of $\boldsymbol{f}_n^p$s and output feature from backbone forms the final resulting feature of the whole model.
This feature can represent the original image for an efficient matching and retrieval of images from the same person in numbers of candidates with unknown identities.

The integration of multiple branches here is a reminiscent of AsNet~\cite{9094042}, which also prompted the diversity of resulting features with asymmetrical branches.
However, our implementation here is much more compact since we propose FPB as the replacement of simple duplication of layer 4 of ResNet in~\cite{9094042}.
Note that we choose ResNet50 as backbone in this work only because its popularity and flexibility with other structures. 
FPB can also be exploited as compatible plugin to other common feature extraction backbones, such as Densenet~\cite{8099726}, InceptionNet~\cite{szegedy2017inception}.

\subsection{Feature Pyramid Branch}
The pivotal structure of FPB is a two layers FPN as shown in Fig.~\ref{fig:architecture}.
This affiliated branch network begins with lateral convolutional filters to convert feature maps with different channel numbers to unified 256.
The lateral filters consist of standard $1\times{1}$ convolutional filters followed by BN and ReLU.
Within the FPB, four low dimensional convolutional filters fullfil the aggregation of features at different scales.
The structures of these filters are similar to lateral filters, except that they adopt $3\times{3}$ kernel.

There exist two cross-scale connections between feature maps with different spatial resolutions in FPB.
One top-down connection is implemented by nearest interpolation to increase the size of feature map.
Conversely, one bottom-up connection is implemented by max pooling with $2\times{2}$ kernel.
There also exist two down-sampling operations as extra edges from input to output node at each layer.
As illustrated by curved arrows in Fig.~\ref{fig:architecture}, we implement down-sampling analogously to the residual structure in ResNet.
As the output of FPB, we take the feature at the deeper layer of FPB and recover the channel to 1024 for consequent processing.
Hence, each layer within FPB can actually be treated as a small bottleneck structure, which extracts information by filters with relatively larger kernel as well as fewer channels.

The design of FPB is inspired by the so-called BiFPN in~\cite{9156454}, which was originally proposed for the task of object detection.
Differing from conventional FPN~\cite{8099589,8579011}, BiFPN proposed to add an extra edge from the input to output node at the same scale.
Our empirical results also prove the efficacy and simplicity of this structure.
However, note that there exist two essential differences between the structures of FPB and BiFPN.
First, FPN aims to tackle the problem of occurring objects with different sizes by aggregating features at different scales, while FPB aims to integrate diverse features from different scales into final matching.
Therefore FPB aggregates features at a single output to average pooling operation, rather than multiple outputs at each layer as BiFPN.
Second, we implement inner nodes as well as down-sampling connections with wider filters than BiFPN.
This is because, the task of Re-ID requires complicated information from different scales for the classification over a relatively larger amount of identities.
In contrast, object detection focuses on a classification problem only over tens of categories, taking COCO detection datasets~\cite{978-3-319-10602-1_48} as a typical instance.
Thus filters with more channels here are adopted to fit the complexity of problem.
Moreover, there also exist some other subtle modifications in FPB according to our empirical results.
For example, we found that weighted feature fusion with learnable parameters at each node is inappropriate in Re-ID.
More detailed analysis can be referred to the ablation study in Section~\ref{sec:ablation}.

\subsection{Self-Attention}% in contrast to local convolution operation, attention take global co-information into account at shallow levels.
\begin{figure}
  % \vskip 0.2in
  \centering
  \subfloat[]{\includegraphics[width=\columnwidth]{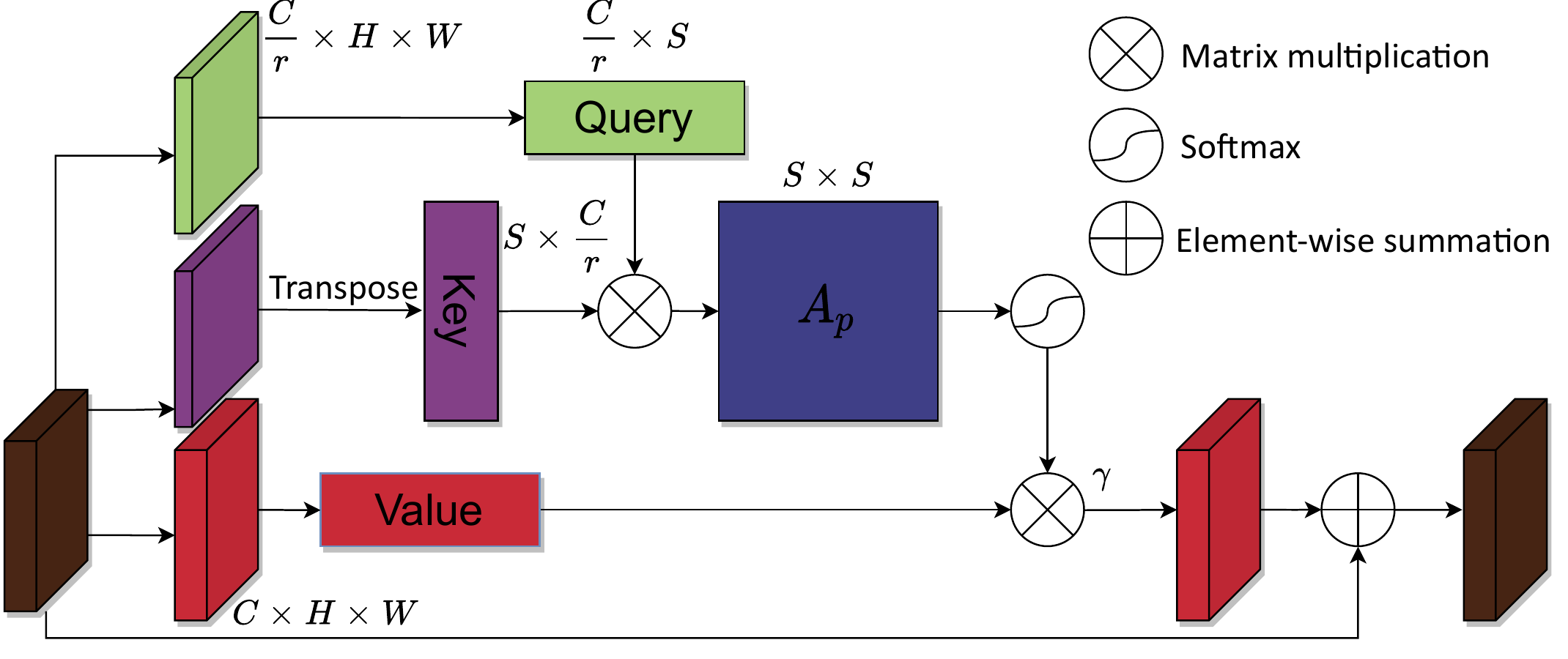}}\label{fig:a_pam} \\
  \subfloat[]{\includegraphics[width=\columnwidth]{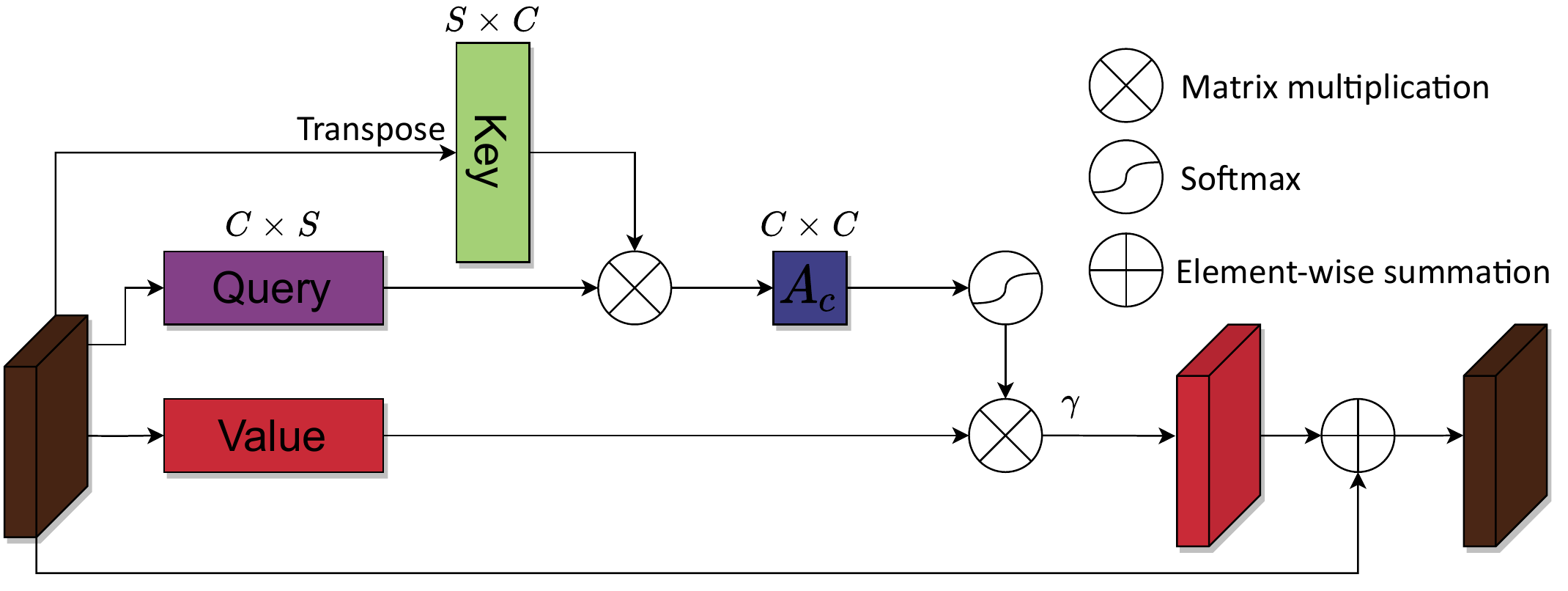}}\label{fig:a_cam}
  \caption{The structures of (a) Position Attention Module and (b) Channel Attention Module.}
  \label{fig:attention}
  % \vskip -0.2in
  \end{figure}

Attention module has been proven as an effective mechanism in various machine learning scenarios.
It exploits the correlation between features to help model focus on more related features and reduce the ambiguity of features for final tasks.
In the case of person Re-ID, the main goal is to train the model to discriminate persons with samples in training set, and to extract representative features from images of unknown identities in testing set for final matching.
This divergence between training and inference requires the model to extract more generic features rather than baised features related to specific instances in training set.
The exploitation of self-attention modules can effectively prompt the generalization capacity of model by forcing it to focus on the relationship of features.
Based on this observation, we insert two self-attention modules at backbone and feature pyramid branch respectively.
Both self-attention modules consist of a PAM followed by a CAM.
Their structures are shown in Fig.~\ref{fig:attention}. 

\textbf{Position Attention Module:} Our implementation of PAM is analogous to ABD-Net~\cite{chen2019ABD}, which can be treated as a simplification of the extensively applied multi-head attention mechanism in Natural Language Processing (NLP)~\cite{vaswani2017attention}.
Given input feature maps $\boldsymbol{X}\in\mathbb{R}^{C\times({H}\times{W})}$, where $C$, $H$ and $W$ are channel number, height and width, respectively.
The PAM projects and reshapes feature at every position onto two lower dimensional subspaces, resulting in query $\boldsymbol{Q}\in\mathbb{R}^{\frac{C}{r}\times{S}}$ and key $\boldsymbol{K}\in\mathbb{R}^{\frac{C}{r}\times{S}}$. 
Here $S=H\times{W}$ is the spatial size of feature map and $r$ is a hyper parameter to control the dimension of subspace.
We follow the usual choice and simply set $r$ as 8 for all experiments.
Specifically, the projections from $\boldsymbol{X}$ to $\boldsymbol{Q}$ and $\boldsymbol{K}$ are implemented by 2$D$ convolution with kernel size as $1\times{1}$.
Then the attention of $\boldsymbol{X}$ can be calculated from query $\boldsymbol{Q}$, key $\boldsymbol{K}$ and value $\boldsymbol{V}$ as
\begin{align}
  \label{eq:attention_pam}
  attention_p(\boldsymbol{X})=\boldsymbol{V}\sigma(\boldsymbol{A}_p)=\boldsymbol{V}\sigma(\boldsymbol{Q}^T\boldsymbol{K}),
\end{align}
where $\sigma(\cdot)$ is the Softmax function, value $\boldsymbol{V}\in\mathbb{R}^{C\times{S}}$ is another projection of $\boldsymbol{X}$ with equivalent dimension. 
Note that if we ignore all learnable parameters here, the position affinity matrix $\boldsymbol{A}_p$ can be simplified as a Gram matrix, which can measure the correlation between features at different positions of $\boldsymbol{X}$.
From this perspective, the essential goal of position attention is to reweigh each feature by its correlations with other features.
As shown in Fig.~\ref{fig:attention}(a), we also adopt a residual structure as well as a learnable parameter $\gamma$ to adjust the impact of attention.

\textbf{Channel Attention Module:} Based on similar motivation, we also implement CAM to extract attention over different channels of features in $\boldsymbol{X}$.
As illustrated in Fig.~\ref{fig:attention}(b), without projection, the channel affinity matrix $\boldsymbol{A}_c$ is directly calculated as
\begin{align}
  \label{eq:attention_cam}
  attention_c(\boldsymbol{X})=(\sigma(\boldsymbol{A}_c)\boldsymbol{X})^T=(\sigma(\boldsymbol{X}\boldsymbol{X}^T)\boldsymbol{X})^T.
\end{align}
A learnable parameter is also adopted here to adjust the impact of attention in the final sum operation with original features.
Comparing with the PAM, the implementation of CAM only requires a few parameters and brings explicit improvement to performance.

\subsection{Cross Orthogonality Regularization}
On top of self-attention modules, we further enforce the diversity of features by orthogonality regularization.
As suggested in~\cite{chen2019ABD}, the orthogonality regularization aims to prompt the representative efficiency of features by reducing the feature correlation between different channels.
The influence is especially obvious on features after attention modules.
Given feature map $\boldsymbol{X}\in\mathbb{R}^{C\times{S}}$, conventional hard regularization normally relies on the Singular Value Decomposition (SVD), which is computationally expensive especially for high dimensional features.
A substitute is the soft orthogonality regularization which optimizes the conditional number of $\boldsymbol{X}\boldsymbol{X}^T$ as
\begin{align}
  \label{eq:conditional_number}
  \mathcal{L}_{or}=\beta||\lambda_1(\boldsymbol{X}\boldsymbol{X}^T)-\lambda_2(\boldsymbol{X}\boldsymbol{X}^T)||_2^2,
\end{align}
where $\beta$ is a small constant, $\lambda_1$ and $\lambda_2$ denote the largest and smallest eigenvalues of matrix, respectively.
With the fast iterative algorithm for solving eigenvalues, the orthogonality regularization can be implemented efficiently during training.

Rather than merely applying orthogonality regularization on single feature map, we propose the Cross Orthogonality Regularization (COR) as shown in Fig.~\ref{fig:architecture}.
Two feature maps, $\boldsymbol{f}^h$ and $\boldsymbol{f}^l$ after attention modules are taken into account jointly here.
Feature maps with different resolutions are unified by max pooling operation, and then concatenated into one higher dimensional feature map.
The orthogonality regularization is then applied to enforce the orthogonality over all channels of features from different positions.
The motivation of COR instead of single orthogonality regularization on different feature maps respectively is that we observe standard back propagation can naturally reduce the correlation over features.
It intrinsically ensures the efficiency of deep models.
In this case, the effectiveness of simple orthogonality regularization on feature map is partially overlapped by the learning of entire model, while COR can be more complementary since it affects multiple branches simultaneously.
Without any extra computation increase to inference, COR brings small by obvious improvement to final performance.

\subsection{Loss Functions}
\label{sec:loss}
At the training stage, we can get four types of output features as shown in Fig.~\ref{fig:architecture}, $\boldsymbol{f}^g$, $\boldsymbol{f}^b$, $\boldsymbol{f}_n^p$s and $\boldsymbol{f}_n^s$s.
Here notation $\boldsymbol{f}_n^p$s and $\boldsymbol{f}_n^s$s mean multiple features from each part after average pooling.
The following loss function is optimized to learn all parameters within the model:
\begin{align}
  \label{eq:loss}
  \mathcal{L}_{total} %=& \mathcal{L}_{id}+\lambda\mathcal{L}_{triplet}\\\nonumber
                      =&\ \alpha\mathcal{L}_{triplet}(\boldsymbol{f}_i^g\odot\boldsymbol{f}_{in}^ps,y_i,\boldsymbol{f}_j^g\odot\boldsymbol{f}_{jn}^ps,y_j)\\\nonumber
                       &+\mathcal{L}_{ce}(\boldsymbol{W}^b\boldsymbol{f}_i^b, y_i)+\sum_{n=1}^N\mathcal{L}_{ce}(\boldsymbol{W}_n^s\boldsymbol{f}_{in}^s, y_i)\\\nonumber
                       &+\mathcal{L}_{cor},
                       %& +\lambda[m+\max_{\boldsymbol{f}_p^o\in{P}(i)}d(\boldsymbol{f}_i^o,\boldsymbol{f}_p^o)-\min_{\boldsymbol{f}_n^o\in{N}(i)}d(\boldsymbol{f}_i^o,\boldsymbol{f}_n^o)]_+
  \end{align}
where $i$ is the index of training sample $(\boldsymbol{x}_i, y_i)$.
$\mathcal{L}_{triplet}(\cdot)$ is the hard mining triplet loss~\cite{hermans2017defense} between sample $i$ and another sample $j$ within a batch.
$\odot$ represents the concatenation operation of all vectors in $\boldsymbol{f}^g$ and $\boldsymbol{f}_n^p$s.
$\mathcal{L}_{ce}(\cdot)$ is the cross entropy loss, $\boldsymbol{W}^b$ and $\boldsymbol{W}_n^s$ are the FC layers after $\boldsymbol{f}^b$ and each $\boldsymbol{f}_n^s$, respectively.
$\mathcal{L}_{cor}$ is the COR version of $\mathcal{L}_{or}$ in Equation~\ref{eq:conditional_number}. 
A hyperparameter $\alpha$ is adopted as a balance between different losses.
The utilization of $\mathcal{L}_{ce}(\cdot)$ follows the conventional framework of learning person Re-ID models~\cite{zheng2016person,luo2019bag}, while the $\mathcal{L}_{triplet}(\cdot)$ prompts the generalization capacity of model by ensuring a larger distance between output features $\boldsymbol{f}^g\odot\boldsymbol{f}_{n}^ps$ from samples of different identities than the same one.

As listed in Equation~\ref{eq:loss}, two kinds of features $\boldsymbol{f}^g$ and $\boldsymbol{f}^b$, delivered by the BNNeck~\cite{luo2019bag} mechanism on the global branch, are included in the loss function.
The BNNeck deploys a BN layer after $\boldsymbol{f}^g$ to get the normalized version $\boldsymbol{f}^b$.
$\boldsymbol{f}^b$ is utilized for the opization of the $\mathcal{L}_{ce}(\cdot)$ on global branch during training as well as part of the final feature during inference.
While original $\boldsymbol{f}^g$ is utilized as part of the output feature to optimize the $\mathcal{L}_{triplet}(\cdot)$ during training.
Our experimental results demonstrate that merely deploying the BNNeck on the global branch rather than both branches can deliver the optimal performance since it naturally forces an asymmetrical structure over outputs and thus ensures the diversity of features from different paths.
% As shown in Fig.~\ref{fig:architecture}, a BNNeck~\cite{luo2019bag} is also exploited here on global branch.
% It deploys a BN layer after $\boldsymbol{f}^g$ to get a normalized version $\boldsymbol{f}^b$ of global feature.
% As listed in Equation~\ref{eq:loss}, $\boldsymbol{f}^b$ is utilized in two cases, output feature for the opization of the classification loss on global branch during training as well as part of the final feature during inference.
% Conversely, original $\boldsymbol{f}^g$ is utilized as part of the output feature to optimize the triplet loss during training.

\section{Empirical Results}
\subsection{Datasets}
In this section, we conduct a series of experiments to analyze the performance of our proposed FPB and compare it with other state-of-the-art works.
Four prevailing person Re-ID datasets are considered here, Market1501~\cite{zheng2015scalable}, DukeMTMC~\cite{Ristani2016Performance} CUHK03~\cite{Li2014DeepReID} and MSMT17~\cite{8578114}.

\textbf{Market1501}~\cite{zheng2015scalable} consists of 32,668 images from 1501 identities captured by six cameras, in which each identity is at least captured by two cameras with multiple images.
For the training set, 12,936 images from 751 identities are considered, leading to an average of 17.2 training samples for one person are available.
For the testing set, 19,732 images from 750 other identities are considered, in which 3,368 images are used as probe set while the rest are used as gallery set.

\textbf{DukeMTMC}~\cite{Ristani2016Performance} consists of 36,411 images from 1,404 identities captured by more than two cameras, and 408 identities captured by only one camera as distractors.
For the training set, 16,522 images from 702 identities are considered.
For the testing set, 17,661 images from 702 other identities are considered, in which 2,228 images are used as probe set while the rest images from the 702 identities as well as distractors are used as gallery set.

\textbf{CUHK03}~\cite{Li2014DeepReID} consists of images from 1467 identities captured by five cameras, in which 767 identities are used as training set and 700 other identities are used as testing set.
The dataset contains two tasks, person Re-ID with labeled images and with detected images.
The labeled dataset has 7,368 images for training and 6,728 images for testing.
The detected dataset has 7,365 images for training and 7,732 images for testing.

\textbf{MSMT17}~\cite{8578114} is relatively larger than aforementioned three datasets.
It consists of 126,441 images from 4,101 identities captured by a 15-camera network (12 outdoor, 3 indoor).
For the training set, 32,621 images from 1,041 identities are considered.
For the testing set, 93,820 images from 3,060 other identities are considered, in which 11,659 images are used as probe set while the rest are used as gallery set.
\begin{table*}
  \centering
  \caption{Comparison of our proposed method with state-of-the-art methods on Market1501, DukeMTMC, CUHK03 (Labeled) and CUHK03 (Detected). The best performances are highlighted by bold font.}
  \label{tab:sota}
  \small
  \setlength{\tabcolsep}{3pt}
  \begin{tabular}{c|>{\centering}p{1.5cm}>{\centering}p{1.5cm}|>{\centering}p{1.5cm}>{\centering}p{1.5cm}|>{\centering}p{1.5cm}>{\centering}p{1.5cm}|>{\centering}p{1.5cm}>{\centering}p{1.5cm}}%p{40pt}m{40pt}|p{40pt}p{40pt}|p{40pt}p{40pt}|p{40pt}p{40pt}|}
  \hline
  \multirow{2}{*}{Method} & \multicolumn{2}{c|}{Market1501} & \multicolumn{2}{c|}{DukeMTMC} & \multicolumn{2}{c|}{CUHK03-Labeled} & \multicolumn{2}{c}{CUHK03-Detected} \tabularnewline
  \cline{2-9}
   & mAP(\%) & rank-1(\%) & mAP(\%) & rank-1(\%) & mAP(\%) & rank-1(\%) & mAP(\%) & rank-1(\%) \tabularnewline
  \hline
  KPM~\cite{shen2018end} & 75.3 & 90.1 & 63.2 & 80.3 & - & - & - & - \tabularnewline
  PCB+RPP~\cite{sun2018beyond} & 81.6 & 93.8 & 69.2 & 83.3 & - & - & 57.5 & 63.7 \tabularnewline
  Mancs~\cite{wang2018mancs} & 82.3 & 93.1 & 71.8 & 84.9 & 63.9 & 69.0 & 60.5 & 65.5 \tabularnewline
  MGN~\cite{wang2018MGN} & 86.9 & 95.7 & 78.4 & 88.7 & 67.4 & 68 & 66.0 & 68.0 \tabularnewline
  MHN~\cite{chen2019mixed} & 85.0 & 95.1 & 77.2 & 89.1 & 72.4 & 77.2 & 65.4 & 71.7 \tabularnewline
  CAMA~\cite{yang2019CAMA} & 84.5 & 94.7 & 72.9 & 85.8 & 66.5 & 70.1 & 64.2 & 66.6 \tabularnewline 
  Bag-Of-Tricks~\cite{luo2019bag} & 85.9 & 94.5 & 76.4 & 86.4 & - & - & - & - \tabularnewline 
  ABD-Net~\cite{chen2019ABD} & 88.28 & 95.6 & 78.59 & 89.0 & - & - & - & - \tabularnewline 
  BDB~\cite{dai2019BDB} & 86.7 & 95.3 & 76.0 & 89.0 & 76.7 & 79.4 & 73.5 & 76.4 \tabularnewline
  Pyramid~\cite{zheng2019pyramidal} & 88.2 & 95.7 & 79.0 & 89.0 & 76.9 & 78.9 & 74.8 & 78.9 \tabularnewline
  SONA~\cite{xia2019SONA} & 88.67 & 95.68 & 78.05 & 89.25 & 79.23 & 81.85 & 76.35 & 79.1 \tabularnewline
  AsNet~\cite{9094042} & 89.13 & 95.50 & 81.10 & 89.93 & 80.87 & 83.23 & 77.17 & 81.43 \tabularnewline  
  Adaptive L2~\cite{ni2020adaptive} & 88.9 & 95.6 & 81.0 & 90.2 & - & - & - & - \tabularnewline  
  % PLR-OSNet~\cite{xie2020learning} & 88.9 & 95.6 & 81.2 & 91.6 & 80.5 & 84.6 & 77.2 & 80.4 \tabularnewline
  RGA-SC~\cite{9157488} & 88.4 & 96.1 & - & - & 77.4 & 81.1 & 74.5 & 79.6 \tabularnewline
  CDNet~\cite{li2021combined} & 86.0 & 95.1 & 76.8 & 88.6 & - & - & - & - \tabularnewline 
  L3DS~\cite{chen2021learning} & 87.3 & 95.0 & 76.1 & 88.2 & - & - & - & - \tabularnewline 
  PAT~\cite{li2021diverse} & 88.0 & 95.4 & 78.2 & 88.8 & - & - & - & - \tabularnewline 
  HOReID~\cite{9351775} & 90.00 & \textbf{96.45} & 81.03 & 89.12 & - & - & - \tabularnewline   
  % SCSN~\cite{9156982} & 88.5 & 95.7 & 79.0 & 91.0 & 84.0 & 86.8 & 81.0 & 84.7 \tabularnewline   
  \hline
  FPB & \textbf{90.6} & 96.1 & \textbf{82.9} & \textbf{91.2} & \textbf{83.8} & \textbf{85.9} & \textbf{81.0} & \textbf{83.8} \tabularnewline   
  \hline
  \end{tabular}
  \end{table*}

\subsection{Implementation Details}
Since we choose ResNet as the backbone, the training framework of our proposed model mainly follows conventional person Re-ID methods~\cite{zheng2016person,sun2018beyond}.
First, the training starts with the pre-trained backbone from ImageNet~\cite{he2016deep}.
For the rest part of the architecture in Fig.~\ref{fig:architecture}, we adopt the widely applied He method~\cite{7410480} as initialization.
Standard augmentation methods including random horizontal flip, random crop, random erasing~\cite{zhong2020random} and random patch~\cite{9011001} are also adopted during training.
We fine-tune the model with Adam optimizer for 120 epochs.
The linear warmup strategy~\cite{luo2019bag} is used, in which the learning rate is initialized at 3.5e-5 and increased to 3.5e-4 in 20 epochs.
Then the learning rate is decayed after 60 and 90 epochs with a rate of 0.1, respectively.

For Market1501, DukeMTMC and CUHK03, the size of input image is resized to $384\times{128}$.
Experiments are executed with a hardware environment as Intel E5-2680CPU at 2.4GHz and a single NVidia Tesla P40 GPU.
The model is trained with a batch size of 64 from 16 identities.
For MSMT17, to prompt the efficiency of learning this massive dataset, we adopt multiple GPUs and keep the batch size at single GPU still to 64.

\textbf{Evaluation Protocol:} For quantitative comparison over different methods, we consider the Cumulative Matching Characteristics (CMC) at mean Average
Precision (mAP) and top-1 accuracy (rank-1) as standard metrics.
All results are obtained without any re-ranking~\cite{zhong2017re} or multi-query fusion~\cite{bai2017reid} techniques.

\subsection{Comparison with State-of-the-art Methods}
In this section, we compare our proposed approach with other state-of-the-art methods. 
In Table~\ref{tab:sota}, we list the performance of different methods on four tasks, Market1501, DukeMTMC, CUHK03 (Labeled) and CUHK03 (Detected).
One can see that our proposed scheme outperforms other approaches with obvious margins.
The only exception is the rank-1 accuracy of HOReID~\cite{9351775} at Market1501.
Here we adopt the result of HOReID~\cite{9351775} with ResNet50 as backbone and an extra Global Hard Identity Searching (GHIS)~\cite{zhang2019learning} method during training.
Without this augmentation, the rank-1 accuracy of HOReID reduces to 95.74\%, which is worse than FPB.
For the mAP, FPB substantially exceeds the second best approaches by 0.6\%, 1.8\%, 2.9\% and 3.8\%.
From our observations in experiments, we treat mAP as a more reliable indicator in scenarios of person Re-ID.
This observation also agrees with~\cite{chen2019ABD}.
Note that the identical architecture with a number of 27.04M learnable parameters is used for all experiments on different datasets.
It only brings an increase of less than 1.5M extra parameters to Resnet50 baseline, which demonstrates the significant efficacy of FPB on extracting representative features.
\begin{table}
  \centering
  \caption{Comparison of our proposed method with state-of-the-art methods on MSMT17. The best performances on different backbones are highlighted by bold font.}
  \label{tab:sota_msmt}
  \small
  \setlength{\tabcolsep}{3pt}
  \begin{tabular}{c|>{\centering}p{1.5cm}|>{\centering}p{1.5cm}>{\centering}p{1.5cm}>{\centering}p{1.5cm}>{\centering}p{1.5cm}}%p{40pt}m{40pt}|p{40pt}p{40pt}|p{40pt}p{40pt}|p{40pt}p{40pt}|}
    \hline
    Method & Backbone & mAP(\%) & rank-1(\%) & rank-5(\%) \tabularnewline
    \hline
    PDC~\cite{8237689} & GoogLeNet & 29.7 & 58.0 & 73.6 \tabularnewline
    GLAD~\cite{wei2017glad} & ResNet50 & 34.0 & 61.4 & 76.8 \tabularnewline
    IANet~\cite{8954262} & ResNet50 & 46.8 & 75.5 & 85.5 \tabularnewline
    BFE~\cite{8954262} & ResNet50 & 51.5 & 78.8 & 89.1 \tabularnewline
    DGNet~\cite{8954292} & ResNet50 & 52.3 & 77.2 & 87.4 \tabularnewline
    OSNet~\cite{9011001} & OSNet & 52.9 & 78.7 & - \tabularnewline
    CDNet~\cite{li2021combined} & CDNet & 54.7 & 78.9 & - \tabularnewline
    ABD-Net~\cite{chen2019ABD} & ResNet50 & 60.8 & 82.3 & \textbf{90.6} \tabularnewline  
    RGA-SC~\cite{9157488} & ResNet50 & 57.5 & 80.3 & - \tabularnewline
    Adaptive L2~\cite{ni2020adaptive} & ResNet50 & 59.4 & 79.6 & - \tabularnewline
    HOReID~\cite{9351775} & ResNet50 & 52.97 & 76.24 & - \tabularnewline
    MGN+PS~\cite{9124699} & ResNet50 & 62.4 & \textbf{84.0} & - \tabularnewline
    FPB & ResNet50 & \textbf{63.5} & 79.8 & 85.4 \tabularnewline
    \hline
    Adaptive L2~\cite{ni2020adaptive} & ResNet101 & 61.9 & 81.3 & - \tabularnewline
    Adaptive L2~\cite{ni2020adaptive} & ResNet152 & 62.2 & 81.7 & - \tabularnewline
    HOReID~\cite{9351775} & ResNet101 & 54.77 & 78.42 & - \tabularnewline
    FPB & ResNet101 & \textbf{63.6} & \textbf{82.5} & \textbf{90.5} \tabularnewline 
    \hline
  \end{tabular}
\end{table}

\textbf{MSMT17:} For the more challenging large scale person Re-ID dataset MSMT17, we compare different configurations of our proposed FPB with other state-of-the-art methods in Table~\ref{tab:sota_msmt}.
Here we implement two versions of FPB with ResNet50 and ResNet101 as backbones respectively.
From the perspective of mAP, one can see that the ResNet50 based implementation of FPB outperforms other methods with obvious gap.
It even exceeds the second best approach, ResNet152 based Adaptive L2~\cite{ni2020adaptive} with more than 60M parameters, from 62.2\% to 63.5\%.
Meanwhile, ResNet101 based FPB retains a parallel performance here with larger backbone.
However from the perspective of rank-1 accuracy, ResNet50 based FPB is worse than the best performance from ABD-Net~\cite{chen2019ABD}.
This situation is mitigated by the larger ResNet101 backbone.
In this case, the FPB dilivers the best performance of rank-1 accuracy.
Note that, even with ResNet101 backbone, the parameters of our proposed FPB is still comparable to ABD-Net.
\begin{table*}
  \centering
  \caption{Influences of each mechanism in the path from IDE as baseline to our final choices of FPB. The results are reported as mAP(\%)/rank-1(\%).}
  \label{tab:ablation1}
  \small
  \setlength{\tabcolsep}{3pt}
  \begin{tabular}{r|c|cccccccc|c}%p{40pt}m{40pt}|p{40pt}p{40pt}|p{40pt}p{40pt}|p{40pt}p{40pt}|}
    \hline
    Method & IDE & & & & & & & & & FPB \tabularnewline
    \hline
    last stride=1? & & $\checkmark$ & $\checkmark$ & $\checkmark$ & $\checkmark$ & $\checkmark$ & $\checkmark$ & $\checkmark$ & $\checkmark$ & \textbf{$\checkmark$}\tabularnewline
    bnneck? & & & $\checkmark$ & $\checkmark$ & $\checkmark$ & $\checkmark$ & $\checkmark$ & $\checkmark$ & $\checkmark$ & \textbf{$\checkmark$} \tabularnewline
    one-layer FPB? & & & & $\checkmark$ & & & & & & \tabularnewline
    two-layers FPB? & & & & & $\checkmark$ & & $\checkmark$ & $\checkmark$ & $\checkmark$ & \textbf{$\checkmark$} \tabularnewline
    three-layers FPB? & & & & & & $\checkmark$ & & & & \tabularnewline
    down-sampling? & & & & & & & $\checkmark$ & $\checkmark$ & $\checkmark$ & \textbf{$\checkmark$} \tabularnewline
    weighted fusion? & & & & & & & & $\checkmark$ & & \tabularnewline
    256 channels? & & & & $\checkmark$ & $\checkmark$ & $\checkmark$ & $\checkmark$ & $\checkmark$ & & \textbf{$\checkmark$} \tabularnewline
    512 channels? & & & & & & & & & $\checkmark$ & \tabularnewline
    \hline
    Market1501 & 81.1/90.8 & 83.7/92.4 & 83.9/92.7 & 82.1/91.6 & 86.2/94.1 & 86.1/94.2 & 88.2/95.0 & 85.0/94.2 & 88.2/95.1 & \textbf{88.2/95.0} \tabularnewline
    \hline
  \end{tabular}
\end{table*}

\subsection{Ablation Study}
\label{sec:ablation}
We carefully construct the final structure of proposed FPB step by step from the IDE as baseline.
The Influences of each attempt are listed in Table~\ref{tab:ablation1}.
First, we empirically prove the effectiveness of proposed tricks in~\cite{luo2019bag}, then we focus on configurations of the feature pyramid structure here.
Typical FPNs in the frameworks of object detection, e.g., Single Shot Detector (SSD)~\cite{978-3-319-46448-02} and PANet~\cite{8579011}, normally retain a structure with more than three layers.
Therefore we construct a \textit{three-layers} FPB as initialization.
It takes feature maps after layer 1, 2 and 3 of backbone as input.
As listed in Table~\ref{tab:ablation1}, our final comparison demonstrates that FPB with two layers after layer 2 and 3 of backbone can deliver the optimal performance, which implies that too detailed local features from small respective field might not be useful for feature matching in Re-ID.
Due to the variation of view and pose, these features could be too unstable to be fixed in specific positions in resulting $\boldsymbol{f}^p$.
This is a major difference between person Re-ID and object detection, in which detailed edge information is required for localization of object.

We also compared other configurations in Table~\ref{tab:ablation1}.
Here the \textit{down-sampling} mechanism refers to the extra edges from input to output node at each layer in FPB.
This operation also bring obvious improvement based on simple connections.
In contrast to the observation in~\cite{9156454}, the \textit{weighted feature fusion} at each node in FPB causes reduction of both mAP nd top-1 accuracy of models.
At last, we compare different widths of convolutional filters including \textit{256 and 512 channel numbers} in the FPB.
It turns out wider filters with extra computation consumptions can merely brings trivial impact here.
Hence, filters with \textit{256 channels} become to our final choice in subsequent experiments.

\textbf{Self-Attention Module:} Although attention module has been proven as an effective mechanism in various vision tasks.
We observe that only deploying them at carefully selected positions can bring positive influence to the whole model.
In FPB, two self-attention modules are injected into the architecture at positions as illustrated in Fig.~\ref{fig:architecture}.
We list the improvements brought by them in Table~\ref{tab:ablation2}.
The deployment of attention modules after the layer 2 of backbone and the lateral filter at the shallower layer of FPB finally delivers the optimal performance. 
This observation agrees with strategy of attention modules in~\cite{9094042}, although there is only one attention module in AsNet.
The extra attention module at FPB further prompts the performance with trivial increase of parameters, since channel number is severely reduced by lateral filters.
\begin{table}
  \centering
  \caption{The influences of self-attention modules and cross orthogonality regularization.}
  \label{tab:ablation2}
  \small
  \setlength{\tabcolsep}{3pt}
  \begin{tabular}{l|ccc}%p{40pt}m{40pt}|p{40pt}p{40pt}|p{40pt}p{40pt}|p{40pt}p{40pt}|}
    \hline
    Method & mAP(\%) & rank-1(\%) & rank-5(\%) \tabularnewline
    \hline
    baseline & 88.2 & 95.0 & 98.4 \tabularnewline
    + Attention on backbone & 89.3 & 95.2 & 98.3 \tabularnewline
    + Attention on FPB & 90.2 & 95.6 & \textbf{98.7} \tabularnewline
    + COR & \textbf{90.6} & \textbf{96.1} & 98.6 \tabularnewline
    \hline
  \end{tabular}
\end{table}

\begin{figure}
  % \vskip 0.2in
  \centering
  \includegraphics[width=\columnwidth]{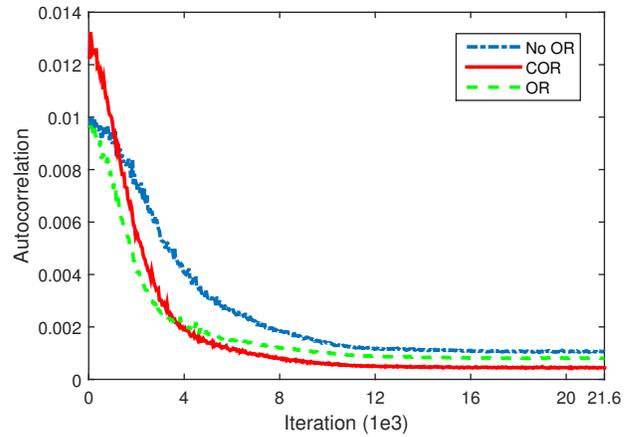}
  \caption{Learning curves of $\mathcal{L}_{or}$s with different configurations of orthogonality regularization.}
  \label{fig:lors}
  % \vskip -0.2in
  \end{figure}

\textbf{Cross Orthogonality Regularization:}
In Table~\ref{tab:ablation2}, we also list the improvement brought by our proposed COR, which delivers the final performance of FPB.
One can see that this joint constraint on features from two layers brings small but non-trivial improvement to the performance of whole model.
We also illustrate the learning curves of $\mathcal{L}_{or}$s in Fig.~\ref{fig:lors}, which reflect the variation of correlations over feature maps during training.
Here we take feature maps at the two positions in Fig.~\ref{fig:architecture} into account, comparing three different strategies of orthogonality regularization during training.
For the first case (No OR), we simply output the sum of correlations without any optimization of regularization.
Then we adopt conventional OR on the two feature maps respectively.
At last, we deploy our proposed COR on the two feature maps to deliver three learning curves.

Note that without any optimization, the learning curve of correlation can still decline during the training.
To reduce the correlation of resulting feature maps is a natural function of standard back propagation learning.
It results in more efficient model as well as more representative output features.
The OR mechanism accelerates this process, helping the feature correlation reach a lower point more efficiently at the beginning of training.
However, one can see that despite due to different calculation method our proposed COR starts from a higher point than other strategies, it finally delivers the lowest correlation between different channels of resulting feature maps.

This observation also helps to understand why injection of attention modules into relatively shallow levels of networks is more effective.
The forward inference of CNNs naturally reduces the feature correlation layer by layer.
Thus available information in Equations~\ref{eq:attention_pam} and~\ref{eq:attention_cam} for the calculation of attention is getting less and less as well. 
\begin{figure}
  % \vskip 0.2in
  \begin{center}
  \centerline{\includegraphics[width=0.5\textwidth]{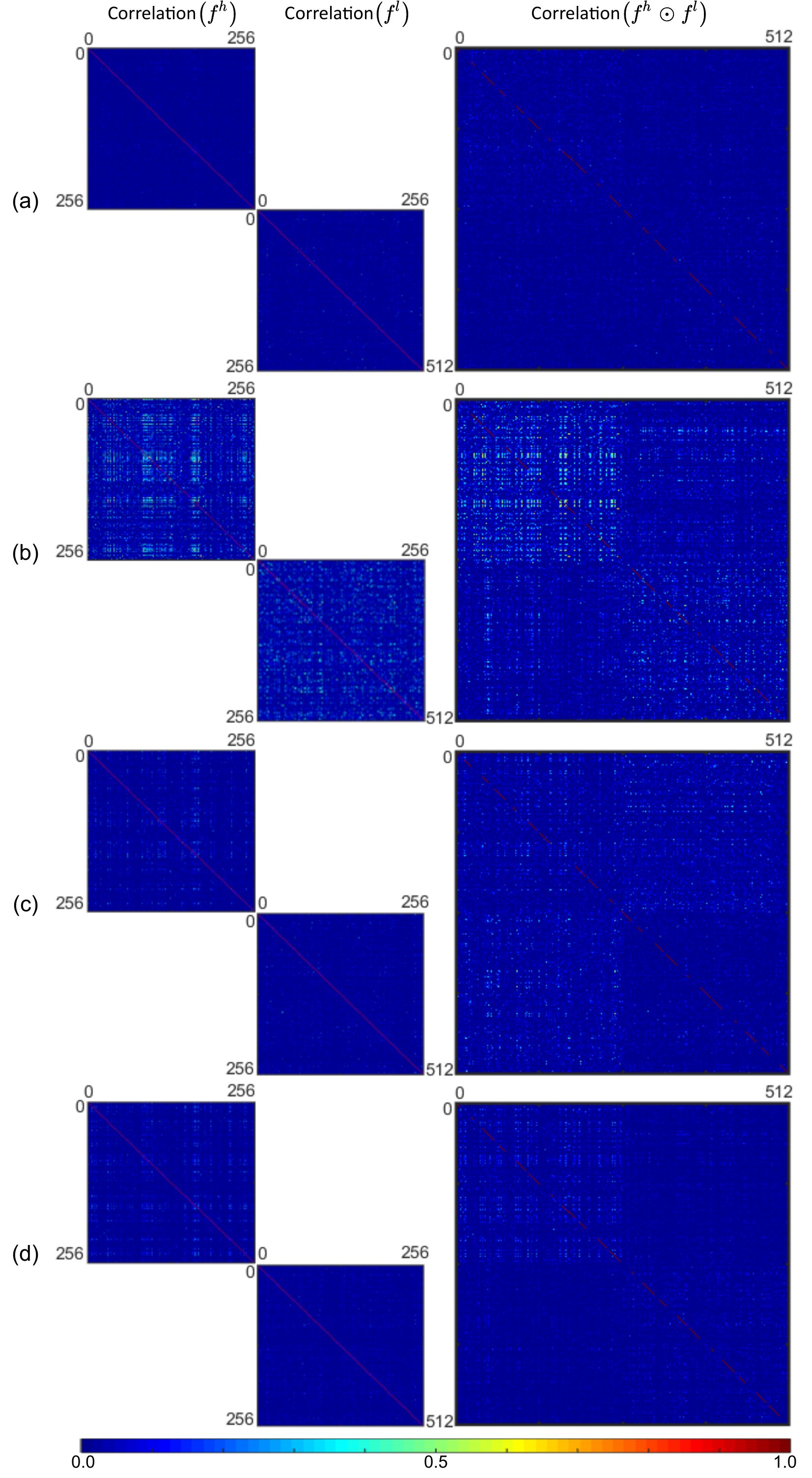}}
  \caption{Visualization of correlation matrices over channels of output feature maps: $\boldsymbol{f}^h$ as the first column, $\boldsymbol{f}^l$ as the second column and concatenation of $\boldsymbol{f}^h$ and $\boldsymbol{f}^l$ ($\boldsymbol{f}^h\odot\boldsymbol{f}^l$) as the third column.
  Here we compare four different configurations: (a) no attention or OR as baseline, (b) attention without OR, (c) attention + OR and (d) attention + COR.
  The relationship between color and correlation is illustrated in the colorbar at the bottom.}
  \label{fig:cor}
  \end{center}
  % \vskip -0.2in
  \end{figure}

From Fig.~\ref{fig:cor}, we can observe this process more inituitively from the correlation of outpuyt feature maps.
From the correlation matrices at the first and second rows, one can see that self-attention modules introduce extra correlation while enhancing the related features for embedding.
This observation agrees with the effect of attention in~\cite{chen2019ABD}.
At Fig.~\ref{fig:cor}(c), the OR mechanism significantly reduces the correlation of both $\boldsymbol{f}^h$ and $\boldsymbol{f}^l$ respectively, improving the efficiency of these features.
However, from the correlation matrix of $\boldsymbol{f}^h\odot\boldsymbol{f}^l$ we can see that correlation between $\boldsymbol{f}^h$ and $\boldsymbol{f}^l$ still exist.
Our proposed COR finally reduces the co-redundancy and further prompts the diversity between different branches of the model.
Another observation from Fig.~\ref{fig:cor}(b) is that correlation of $\boldsymbol{f}^h$ is larger than $\boldsymbol{f}^l$.
This observation also agrees with our aforementioned assumption, the correlation of features is naturally shrinked during inference of CNNs to ensure a more efficient extraction.

\begin{figure}
  % \vskip 0.2in
  \centering
  \subfloat[]{\includegraphics[width=0.5\textwidth]{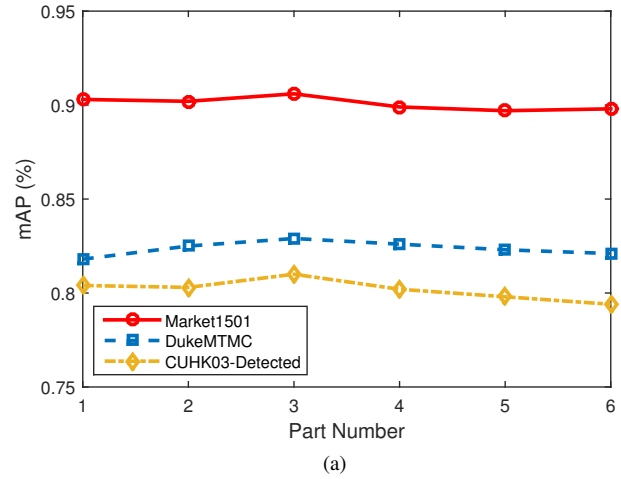}}\label{fig:maps_pn} \\
  \subfloat[]{\includegraphics[width=0.5\textwidth]{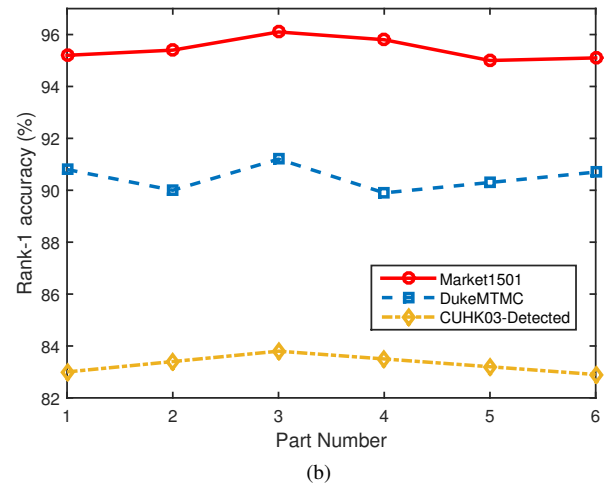}}\label{fig:r1_pn}
  \caption{The (a) mAP and (b) rank-1 vs. part number over three person Re-ID datasets.}
  \label{fig:partnumber}
  % \vskip -0.2in
\end{figure}

\textbf{Influence of Part Number:} For the output features $\boldsymbol{f}^p$ of our proposed FPB, we adopt a similar structure as~\cite{9094042}, in which several features represent corresponding parts of image respectively.
In Fig.~\ref{fig:partnumber}, we study the variations of mAP and top-1 accuracy alongside the change of part numbers as well.
It can be shown that on three different datasets, the tendency is consistent and explicit.
The configuration of three parts in $\boldsymbol{f}^p$ has been proven as optimal and is adopted in all of our experiments.
This configuration is the same as~\cite{9094042}, while differing from six parts in the original part-based framework for person Re-ID~\cite{sun2018beyond}. 
It implies that a smaller part number as part branch can work better with the help from global branch.

\subsection{Visualization}
\label{sec:visual}
\begin{figure}
  % \vskip 0.2in
  \centering
  \includegraphics[width=\columnwidth]{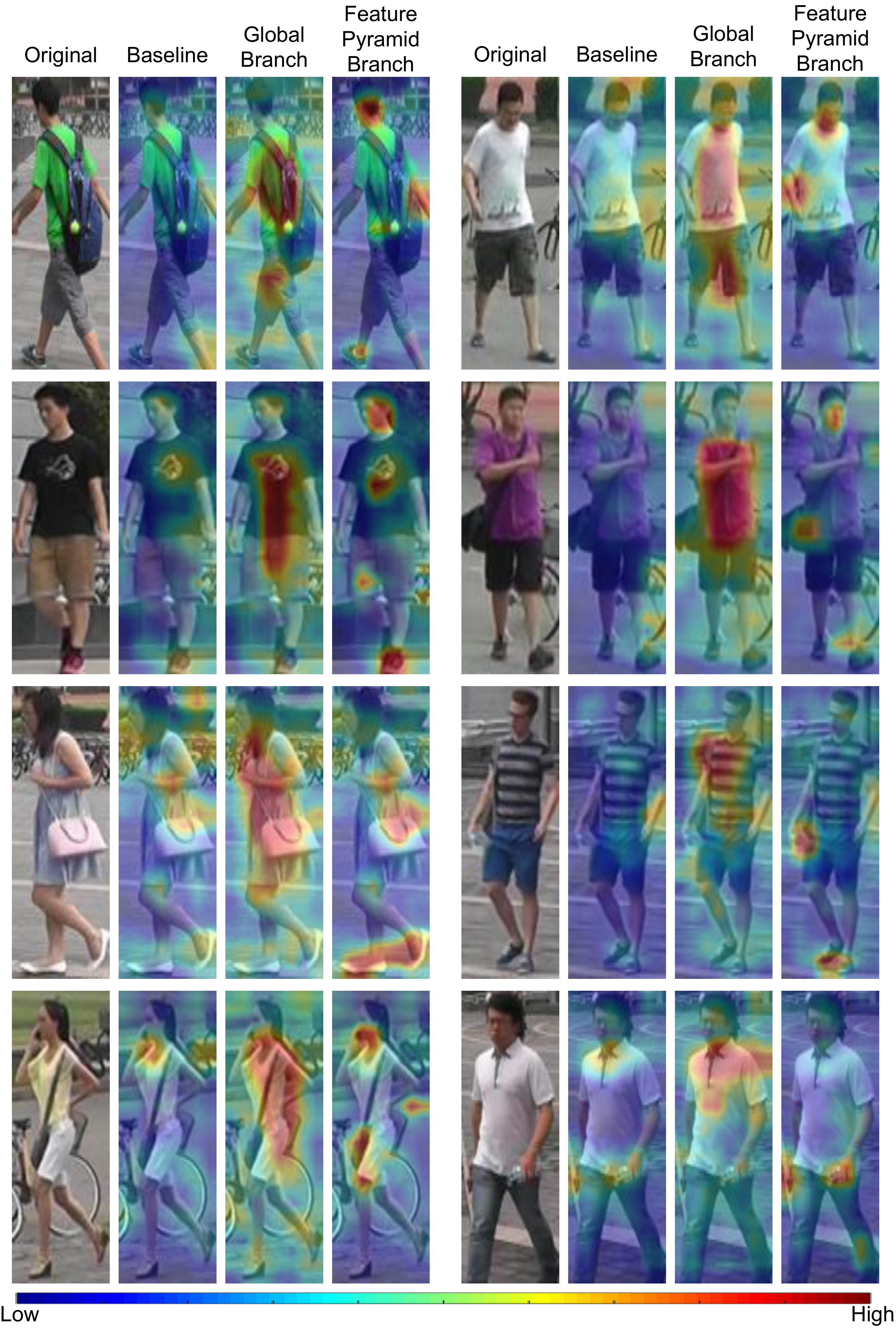}
  \caption{Visualization of samples and corresponding attention maps.
  Columns one and five are original images.
  Columns two and six are activation maps of output of IDE as baseline.
  Others are activation maps of feature maps at different positions of FPB.}
  \label{fig:am}
  % \vskip -0.2in
  \end{figure}

To further study the influence of our proposed dual-branch structure, we illustrate some feature embedding samples with their activation maps~\cite{yang2019CAMA} in Fig.~\ref{fig:am}.
For each input image, we compare activation maps from three different configurations.
The first one is generated by IDE as baseline.
The second one is activation map of output feature map after the layer 4 of global branch in Fig.~\ref{fig:architecture}.
The third one is at the output of the feature pyramid branch before average pooling.

Fig.~\ref{fig:am} shows two observations: (1) Rather than small highlighted points in IDE, the output feature maps of global branch correspond to higher activation over the main part of person.
This is mainly because of the utilization of self-attention modules.
(2) The attentive feature branch serves as complementary function, focusing on more detailed local parts of person.
Such diversity is ensured by two mechanisms.
An asymmetrical architecture between branches naturally lead to different processses of features, help branches focus on different parts during training.
Also, the triplet loss function over the concatenated output features from both branches encourages more diverse information being extracted.

\section{Conclusion and Future Work}\label{sec:conclusion-and-future-work}
We propose a novel structure called feature pyramid branch for the task of Person Re-ID.
The structure can server as an affiliated branch to complement the backbone.
Features at different scales can be extracted and aggregated in this branch for the final embedding with higher diversity.
Cooperated with attention mechanism as well as orthogonality regularization, our proposed FPB structure can significantly prompt the model performance with trivial increase of computational complexity.
For future work, we will continure to investigate the potentiality of the feature pyramid structure in various scenarios of feature embedding, specifically its compatibility with more lightweight backbones.

\ifCLASSOPTIONcaptionsoff
  \newpage
\fi

% trigger a \newpage just before the given reference
% number - used to balance the columns on the last page
% adjust value as needed - may need to be readjusted if
% the document is modified later
%\IEEEtriggeratref{8}
% The "triggered" command can be changed if desired:
%\IEEEtriggercmd{\enlargethispage{-5in}}

% references section

% can use a bibliography generated by BibTeX as a .bbl file
% BibTeX documentation can be easily obtained at:
% http://mirror.ctan.org/biblio/bibtex/contrib/doc/
% The IEEEtran BibTeX style support page is at:
% http://www.michaelshell.org/tex/ieeetran/bibtex/
%\bibliographystyle{IEEEtran}
% argument is your BibTeX string definitions and bibliography database(s)
%\bibliography{IEEEabrv,../bib/paper}
%
% <OR> manually copy in the resultant .bbl file
% set second argument of \begin to the number of references
% (used to reserve space for the reference number labels box)
%\begin{thebibliography}{1}
%
%\bibitem{IEEEhowto:kopka}
%H.~Kopka and P.~W. Daly, \emph{A Guide to \LaTeX}, 3rd~ed.\hskip 1em plus
%  0.5em minus 0.4em\relax Harlow, England: Addison-Wesley, 1999.
%
%\end{thebibliography}
\bibliographystyle{IEEEtran}
\bibliography{egbib}
% biography section
% 
% If you have an EPS/PDF photo (graphicx package needed) extra braces are
% needed around the contents of the optional argument to biography to prevent
% the LaTeX parser from getting confused when it sees the complicated
% \includegraphics command within an optional argument. (You could create
% your own custom macro containing the \includegraphics command to make things
% simpler here.)
%\begin{IEEEbiography}[{\includegraphics[width=1in,height=1.25in,clip,keepaspectratio]{mshell}}]{Michael Shell}
% or if you just want to reserve a space for a photo:

%\begin{IEEEbiography}{Michael Shell}
%Biography text here.
%\end{IEEEbiography}
%
%% if you will not have a photo at all:
%\begin{IEEEbiographynophoto}{John Doe}
%Biography text here.
%\end{IEEEbiographynophoto}
%
%% insert where needed to balance the two columns on the last page with
%% biographies
%%\newpage
%
%\begin{IEEEbiographynophoto}{Jane Doe}
%Biography text here.
%\end{IEEEbiographynophoto}

% You can push biographies down or up by placing
% a \vfill before or after them. The appropriate
% use of \vfill depends on what kind of text is
% on the last page and whether or not the columns
% are being equalized.

%\vfill

% Can be used to pull up biographies so that the bottom of the last one
% is flush with the other column.
%\enlargethispage{-5in}

% that's all folks
\end{document}